\pdfoutput=1
\documentclass[11pt]{article}

\usepackage[]{acl}

\usepackage{times}
\usepackage{latexsym}

\usepackage[T1]{fontenc}
\usepackage[utf8]{inputenc}

\usepackage{microtype}

\usepackage{microtype}
\usepackage{bbm}
\usepackage{amsmath}
\usepackage{graphicx}
\usepackage{booktabs}
\usepackage{graphicx}
\usepackage{caption}
\usepackage{subcaption}
\usepackage[normalem]{ulem}
\usepackage{multirow}

\useunder{\uline}{\ul}{}

\newcommand{\knnmt}{$k$NNMT\ }
\newcommand{\knnmtend}{$k$NNMT}

\newcommand{\method}{\textsc{PRED}\ }
\newcommand{\methodend}{\textsc{PRED}}

\title{Better Datastore, Better Translation: Generating Datastores from Pre-Trained Models for Nearest Neural Machine Translation}

\author{
    Jiahuan Li$^1$\footnotemark[1], Shanbo Cheng$^2$, Zewei Sun$^2$, Mingxuan Wang$^2$, Shujian Huang$^1$ \\
    $^1$National Key Laboratory for Novel Software Technology, Nanjing University, China \\
    $^2$ByteDance AI Lab \\
    \texttt{$^1$lijh@smail.nju.edu.cn, huangsj@nju.edu.cn} \\
    \texttt{$^2$\{chengshanbo, wangmingxuan.89, sunzewei.v\}@bytedance.com}
}

\begin{document}
\maketitle
\renewcommand{\thefootnote}{\fnsymbol{footnote}}
\footnotetext[1]{Work is done while at ByteDance.}
\renewcommand{\thefootnote}{\arabic{footnote}}
\begin{abstract}

Nearest Neighbor Machine Translation (\textbf{\knnmtend}) is a simple and effective method of augmenting neural machine translation (\textbf{NMT}) with a token-level nearest neighbor retrieval mechanism. The effectiveness of \knnmt directly depends on the quality of retrieved neighbors. However, original \knnmt builds datastores based on representations from NMT models, which would result in poor retrieval accuracy when NMT models are not good enough, leading to sub-optimal translation performance. In this paper, we propose \methodend, a framework that leverages \textbf{Pre}-trained models for \textbf{D}atastores in $k$NN-MT. Better representations from pre-trained models allow us to build datastores of better quality. We also design a novel contrastive alignment objective to mitigate the representation gap between the NMT model and pre-trained models, enabling the NMT model to retrieve from better datastores. We conduct extensive experiments on both bilingual and multilingual translation benchmarks, including WMT17 
English $\leftrightarrow$ Chinese, WMT14 English $\leftrightarrow$ German, IWSLT14 German $\leftrightarrow$ English, and IWSLT14 multilingual datasets. Empirical results demonstrate the effectiveness of \methodend. 
\end{abstract}
\section{Introduction}

Retrieval-Enhanced Neural Machine Translation (\textbf{RE-NMT}) aims to augment parametric neural translation models with an external retrieval module, and has been proven to be effective in many works \citep{gu2018search,bapna-firat-2019-non,knnmt,cai-etal-2021-neural}. Being able to access parallel corpus at the inference time, RE-NMT shows more expressiveness than pure parametric methods.

\knnmt \citep{knnmt} is a representative work of RE-NMT, and has attracted much attention due to its conceptual simplicity and impressive performance \citep{zheng-etal-2021-adaptive,jiang-etal-2021-learning,zheng-etal-2021-non-parametric,fast_knnmt,wang-etal-2022-efficient,CLKNN}. 
To augment an existing NMT model, \knnmt pre-builds a datastore by storing all token-level translation examples as \textit{(key, value)} pairs, where keys are decoding states of the MT model that encodes the source sentence and target sentence prefix, and values are target tokens corresponding to the decoding states. 
When retrieving, the decoding state from the same NMT model is treated as \textit{query} to retrieve $k$ nearest (key, value) pairs based on distances between the query and keys. 
The retrieved values are fused to the given MT model to help with translation.

\begin{figure}
    \centering
    \includegraphics[width=1.0\linewidth]{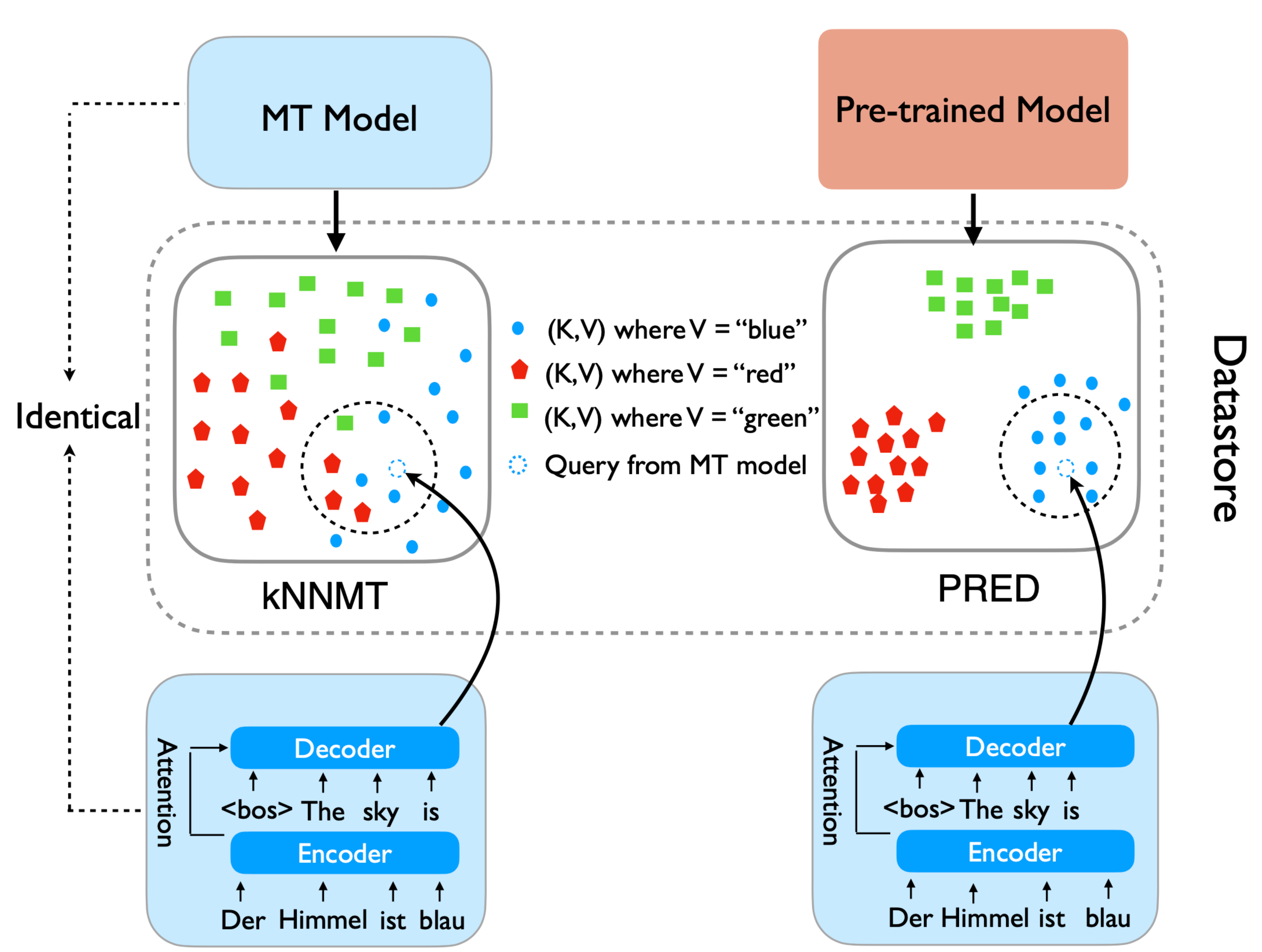}
    \caption{Schematic illustration of the retrieval process in \knnmt and \methodend. Scatters in the figure depict the representation space of datastores in \knnmt and \methodend.
    \textit{Left}: In \knnmtend, the K-V consistency might be low because the NMT model does not always produce good representations. \textit{Right}: 
    \method generates datastores with high K-V consistency leveraging powerful pre-trained models.
    }
    \label{fig:scheme}
\end{figure}

The effectiveness of the retrieval process in \knnmt is strongly affected by Key-Value (K-V) consistency, which quantifies whether similar key representations in the datastore correspond to the same tokens. With low K-V consistency, the neighborhood of a query would contain incorrect value tokens.  Since datastores in \knnmt are built from NMT model, the K-V consistency of the built datastore is heavily limited by the quality of NMT models' representation space, leading to sub-optimal retrieval quality and translation performance \citep{wang-etal-2022-efficient,CLKNN}.

In this paper, we propose \methodend, a framework that leverages pre-trained models (PTMs) for datastore generation in \knnmtend. Compared to vanilla NMT models, PTMs are trained on a much larger scale corpus, which leads to better representations. This enables us to obtain datastores with higher K-V consistency.
To bridge the representation discrepancy between MT models and datastores generated by PTMs, we design a novel contrastive objective to align queries to their corresponding keys, helping the model to retrieve proper neighborhood examples. 
The schematic illustration of \method is shown in Figure \ref{fig:scheme}.

We evaluate our framework in both bilingual and multilingual translation benchmarks, including WMT17 Chinese $\leftrightarrow$ English, WMT14 English $\leftrightarrow$ German, IWSLT14 German $\leftrightarrow$ English, and IWSLT14 multilingual datasets. Extensive experimental results demonstrate the superiority of \method compared to competitive baselines in terms of translation quality. Moreover, compared to using PTMs in the MT models in \knnmtend, \method does not increase the MT model size, thus significantly saving computational and storage costs.

\section{\knnmt and Its Limitation}
\label{sec:background}
\subsection{\knnmt}
 \knnmt augments NMT predictions with retrieved examples from a pre-built datastore. Given a parallel corpus $\mathcal{D} = \{(X_i, Y_i)\} $, and a pre-trained NMT model $f$, \knnmt builds a datastore by collecting all token-level examples  in $\mathcal{D}$, and each example is a (key, value) pair in the form of 
\begin{equation}
    (\mathbf{k},v) = ( f(X,Y_{<t}), y_t),
\end{equation}

\noindent where $f(X,Y_{<t})$ is a contextual representation from MT decoder by teacher forcing decoding on the sentence pair $(X,Y)$, and $y_t$ is the corresponding $t$-th target token.

At the inference time,  \knnmt predicts the target token $y_t$ relying on not only the probability distribution from the NMT model, but also retrieved examples from the datastore. Specifically, given the already generated tokens $Y_{<t}$, the contextual representation $f(X,Y_{<t})$ is computed as query $\mathbf{q}_t$ to retrieve $k$ neighbors from the datastore w.r.t some distance function \footnote{Popular choices are Euclidean distance and cosine distance.} $d(\cdot,\cdot)$. Denoting the retrieved neighbors as $\mathcal{N}=\{(\mathbf{k}_i, v_i), i= 1,2,\ldots, k\}$, the distribution from $k$NN is computed as:
\begin{equation}
    p_{k\text{NN}} (y|X,Y_{<t}) \propto \sum_{(\mathbf{k}_i, v_i) \in \mathcal{N}_q} \mathbbm{1} [y = v_i] \cdot  e ^{ \frac{- d( \mathbf{q}_t, \mathbf{k}_i)}{T}},
\end{equation}

\noindent where $T$ is the temperature hyperparameter. The final probability is an interpolation of MT model probability $p_{\text{MT}}$ and $k$NN probability $p_{k\text{NN}}$:
\begin{align}
\begin{split}
    p(y | X, Y_{<t}) = \lambda  & p_{\text{MT}}(y|X, Y_{<t}) \\
                            & +  (1- \lambda) p_{k\text{NN}}(y | X, Y_{<t}).
\end{split}
\label{eq:knnmt}
\end{align}

\subsection{Analysis On Datastores of \knnmt}
For \knnmtend, it is crucial to retrieve accurate (key, value) pairs to obtain good performance. Intuitively, we identify two factors that affect retrieval quality: query-key consistency and key-value consistency. Q-K consistency 
%measures the similarity between query and key given similar contexts, and 
is quantified by the cosine similarity between queries and their corresponding keys, and 
K-V consistency 
%measures whether similar keys in the datastore corresponds to the same tokens, and 
is computed by treating each key as a query, and computing the proportion of entries that share the same value with the query among $k$ neighbors:
%the accuracy of the leave-one-out $k$ nearest neighbor classifier on all entries in the datastore $\mathcal{S}$: 
\begin{align}
   \text{KV-Cons}(\mathcal{S}) = \frac{1} {|\mathcal{S}| \cdot k} \sum_{\mathbf{q} \in \mathcal{S}} \sum_{(\mathbf{k}_i,v_i) \in \mathcal{N}_q} \mathbbm{1}[y(\mathbf{q}) = v_i],
\label{eqn:kv_cons}
\end{align}

\noindent where $\mathcal{N}_q$ is the set of $k$ neighbors of the query $q$.

 Conceptually the Q-K consistency of \knnmt is perfect. This is because queries and keys in \knnmt are generated by the same NMT model. However, due to the limited size of the parallel corpus, representations from the NMT model tend to be not satisfying enough. This would result in sub-optimal K-V consistency of \knnmtend, i.e. there could be some regions in the representation space  where keys are similar while values are with distinct semantics. When a query is projected to these regions, the retrieval results would be useless, if not harmful, to the translation process.
This motivates us to seek better ways to generate datastores, other than using the original NMT representations.

\section{Method}

In this section, we introduce \methodend, a framework that augments \knnmt with stronger datastores. We start by introducing our unified model architecture that performs translation and retrieval simultaneously. We then propose an auxiliary training objective to bridge the representation gap between the translation model and external datastores. Finally, we describe how to make use of stronger machine translation models or pre-trained language models to build datastores.
\begin{figure}[t]
    \centering
    \includegraphics[width=1.0\linewidth]{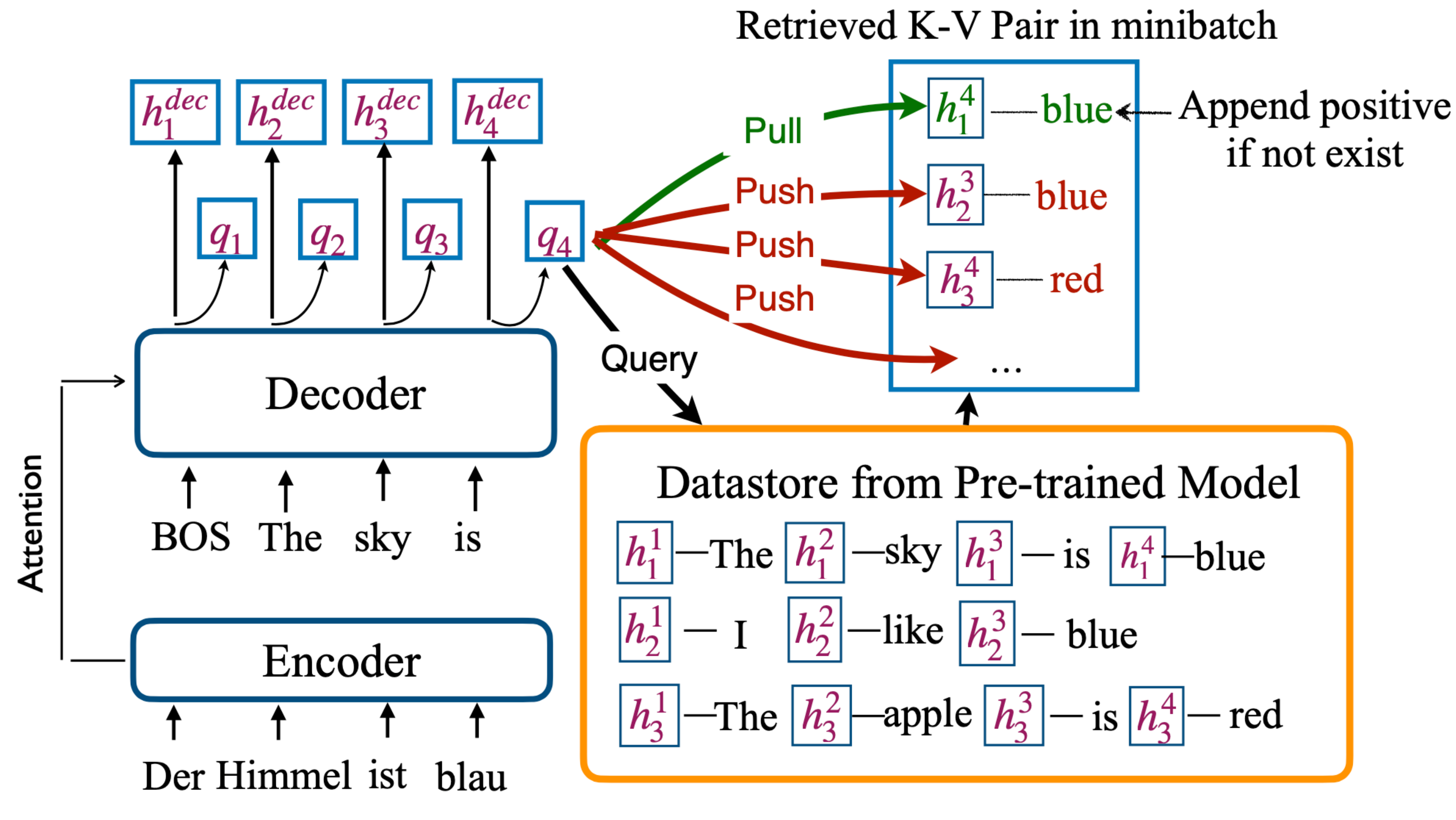}
    \caption{The training paradigm of our methods. Translation of the German sentence is ``The sky is blue".}
    \label{fig:arch}
\end{figure}
\subsection{Model Architecture}

Our model is based on Transformer \citep{transformer} architecture. Given an input sentence $X$ and the translation history $Y_{<t}$, we extract two representations from the Transformer decoder: translation state $\mathbf{h}_t$ and retrieval query $\mathbf{q}_t$, where $\mathbf{h}_t$ is the output of the final decoder layer, and $\mathbf{q}_t$ is the input representation to the feed-forward network in the last decoder layer following \knnmt \citep{knnmt}.

Given query $\mathbf{q}_t$, we retrieve $k$ neighbors from the datastore $\mathcal{D}$\footnote{Ways to create datastores will be described in Section \ref{datastore_creation}.}, which is denoted as $(\mathbf{k}_1, v_1), (\mathbf{k}_2, v_2), ..., (\mathbf{k}_k, v_k)$. Following \citet{yogatama-etal-2021-adaptive}, we integrate the retrieval results at the representation level instead of the probability level. This avoids manually tuning the mixing weight $\lambda$ and temperature $T$ in the original \knnmtend. Concretely, we embed each $v_i$ as $\mathbf{e}_i$ using the target-side word embedding matrix $\mathbf{W}_e$. Then we use a simple attention mechanism to aggregate $\mathbf{e}_1, \mathbf{e}_2, ..., \mathbf{e}_k$ to a single vector $\mathbf{m}$, which is fused to the translation state in a gated fashion:
\begin{align}
% \begin{split}
\mathbf{m}_t &= \sum_{j=1}^{j} \frac{ \text{exp} ( \mathbf{q}_t^T \mathbf{e}_j)} {\sum_{i=1}^{K} \text{exp} (\mathbf{q}_t^T \mathbf{e}_i ) } \cdot \mathbf{e}_j\\
\mathbf{g}_t &= \sigma( \mathbf{W}_1 \mathbf{h}_t + \mathbf{W}_2 \mathbf{m}_t + b) \\
\mathbf{z}_t &= \mathbf{g}_t \odot \mathbf{m}_t + (1 - \mathbf{g}_t) \odot \mathbf{h}_t,
% \end{split}
\end{align}

\noindent where $\mathbf{W}_1, \mathbf{W}_2$ are transforming matrices, $b$ is the bias term, and $\sigma$ is the sigmoid function. The final generation probability is computed as 
\begin{equation}
    p(y_t | X, Y_{<t}) = \text{softmax}(\mathbf{z}_t; \mathbf{W}_e).
\end{equation}

\subsection{Learning}

Ideally, the query $\mathbf{q}_t$ should retrieve neighbors that are semantically similar to the target token $y_t$, so that they would be helpful for translation. 
However, since there exists a discrepancy between the representation space of the MT model and the pre-trained model, directly using query $\mathbf{q}_t$ to retrieve from this external datastore would result in meaningless retrieval results.
To alleviate this problem, we propose Neighborhood Contrastive Aligning (NCA), an auxiliary training objective that explicitly aligns the retrieval query representation to its corresponding key representation in the datastore. Figure \ref{fig:arch} illustrates the training process.

Specifically, we treat each query as the anchoring example, and its corresponding key representation as the positive example. The set of negative examples $\mathcal{K}$ consists of all retrieved neighborhood keys in the mini-batch, summing up to $B \times k$ examples, where $B$ is the number of tokens in the mini-batch, and $k$ is the number of retrieved neighbors of each query. We then minimize a contrastive objective as follows:
\begin{equation}
    \mathcal{L}_{\text{NCA}} (X,Y) = - \text{log} \sum_{t=1}^{|Y|} \frac {\text{exp} (\mathbf{q}_{t}^{T} \mathbf{k}_t / \tau)} { \sum_{\hat{\mathbf{k}} \in \mathcal{K}} \text{exp} (\mathbf{q}^T_t \hat{\mathbf{k}} / \tau )} ,
\end{equation}

\noindent where $\mathbf{q}_t$ is the retrieval query at the $t$ timestep, $\mathbf{k}_t$ is the corresponding key representation of $\mathbf{q}_t$, and $\hat{\mathbf{k}}$ is another retrieved key representation in the batch. $\tau$ is the temperature parameter.

\paragraph{Overall Objective} We train our model by optimizing the translation objective and alignment objective simultaneously:
\begin{align}
    \mathcal{L}_{\text{MT}}(X, Y) &= - \sum_{i=1}^{|Y|} \text{log}\  p(y_t|Y,Y_{<t}) \\
    \mathcal{L}_{\text{overall}} &= \mathcal{L}_{\text{MT}} + \alpha \mathcal{L}_{\text{NCA}},
\end{align}

\noindent where $\alpha$ is the hyperparameter that balances the translation objective and alignment objective. 

Unlike previous works \citep{cai-etal-2021-neural}, we do not update representations in the datastore when training since they usually come from powerful pre-trained models, and further finetuning would cause catastrophic forgetting of previously acquired language knowledge.

\subsection{Datastore Creation from Pre-trained Models}
\label{datastore_creation}
In this section, we introduce two ways to build better datastores using pre-trained models: datastore creation from MT models and from  pre-trained language models (PLM).

\subsubsection{MT model based Datastore Creation}
For MT model based datastore creation, we follow the procedure \knnmt used, except that representations come from another more powerful MT model, such as models that are specifically pre-trained for machine translation \citep{lin-etal-2020-pre,pan-etal-2021-contrastive}. 
%Our method enables us to retrieve from better datastores while avoiding computing through the large MT model at inference time.
Our method enables us to leverage these powerful models without paying as much computational cost as using them as NMT initialization.

One may think our method resembles \textit{knowledge distillation} (KD) \citep{kd,kim-rush-2016-sequence}. However, we would like to point out that our method is different from KD in terms of both motivation and technical details. On the one hand, KD aims to learn a lightweight student model under the supervision of a heavy teacher model, while our goal is to enable MT models to retrieve from better datastores. On the other hand, traditionally KD directly minimizes the mean squared error (MSE) between representations of the student model and teacher model, while we propose a novel contrastive objective to align the representation space of NMT models and PLMs. We discuss the empirical influence of training objectives in Section \ref{sec:mse_vs_contrast}.

\begin{table*}[]
\scriptsize
\centering

\begin{tabular}{c|cc|cc|cc|cc|cc|cc}
\toprule[1pt]
            & \multicolumn{4}{c|}{WMT17}   & \multicolumn{4}{c|}{WMT14}   & \multicolumn{4}{c}{IWSLT14} \\
            \midrule
 & 
  \multicolumn{2}{c|}{En $\to$ Zh} &
  \multicolumn{2}{c|}{Zh $\to$ En} &
  \multicolumn{2}{c|}{En $\to$ De} &
  \multicolumn{2}{c|}{De $\to$ En} &
  \multicolumn{2}{c|}{En $\to$ De} &
  \multicolumn{2}{c}{De $\to$ En} \\ 
            & BLEU & COMET & BLEU & COMET & BLEU & COMET & BLEU & COMET & BLEU & COMET & BLEU & COMET \\
            \midrule
Transformer &34.0 & 53.4 & 23.5 & 43.5 & 27.3 & 43.9  & 31.5 & 33.5    & 28.2 & 26.7  & 34.5 & 23.3  \\
\knnmt    &  34.5 & 55.5 & 24.1 & 47.3 & 27.6 & 45.7  & 31.6 & 35.2   & 28.5 & 27.8  & 34.8 & 24.6  \\
\midrule
\multicolumn{13}{l}{\textit{w/ XLM-R (base)} \citep{conneau-etal-2020-unsupervised}}                                                          \\
\midrule
Transformer $^\dag$ & 34.9 & 56.1 & 24.1 & 47.8 & 27.8 & 47.8  & 32.5 & 36.9    & 29.3 & 31.4  & 35.3 & 33.7  \\
\knnmt $^\dag$  &   35.2 & 56.8 & 24.1 & 47.2 & 27.9 & 49.0  & 32.7 & 37.8   & 29.6 & 32.1  & 35.5 & 34.4  \\
\method $^\dag$  &  \textbf{35.5} & \textbf{57.2} & \textbf{24.6} & \textbf{47.9} & \textbf{28.7} & \textbf{51.3}  & \textbf{33.1} & \textbf{38.5}   & \textbf{30.1} & \textbf{33.4}  & \textbf{36.4} & \textbf{34.9}  \\
\midrule
\multicolumn{13}{l}{\textit{w/ mRASP2} \citep{pan-etal-2021-contrastive}}                                                               \\
\midrule
Transformer $^\ddag$ & 34.4 & 55.0 & 23.9 & 46.1 & 28.1 & 47.6  & 32.7 & 36.4   & 29.9 & 32.1  & 35.4 & 33.7  \\
\knnmt $^\ddag$  &  34.8 & 55.3 & 24.2 & 46.9 & 28.3 & 48.0  & 32.9 & \textbf{38.2}    & \textbf{30.0} & 31.6  & 35.7 & 34.0  \\
\method $^\ddag$  & \textbf{35.0} & \textbf{56.9} & \textbf{24.4} & \textbf{47.4}  & \textbf{28.8} & \textbf{51.5}  & \textbf{33.1} & \textbf{38.2}  & \textbf{30.0} & \textbf{32.4}  & \textbf{36.2} & \textbf{35.2} \\
\bottomrule[1pt]
\end{tabular}
\caption{Bilingual translation results on WMT17 En$\ \leftrightarrow \ $Zh, WMT14 En$\ \leftrightarrow \ $De and IWSLT14 De$\ \leftrightarrow \ $En datasets. Translation quality are evaluated by BLEU [\%] and COMET [\%]. Best performances across each setting are in bold. $\dag$: Models that take XLM-R's representations as encoder input. $\ddag$: Models that are trained with knowledge distillation objective from mRASP2. }
\label{tab:main_results}
\end{table*}

\subsubsection{PLM based Datastore Creation}
Another viable choice is to utilize large-scale pre-trained language models for datastore creation. Existing PLMs can be divided into three categories according to training objectives: \textbf{(1)} masked language models (MLM) \textbf{(2)} denoising autoencoders (DAE) 
\textbf{(3)} causal language models (CLM). We design different strategies for each kind of model, respectively.

% \paragraph{Datastore Creation for MLM models}
For MLM models $f_{\text{MLM}}$, e.g. BERT \citep{devlin-etal-2019-bert}, RoBERTA \citep{roberta}, XLM-R \citep{conneau-etal-2020-unsupervised}, we feed target-side sentences to them, and obtain the contextual representation at the top layer as keys, with the input tokens as values:
\begin{align}
\begin{split}
    \mathbf{H} &= f_{\text{MLM}}(Y) \\
    (k,v) &= (\mathbf{H}_t, y_t).
\end{split}
\end{align}

% \paragraph{Datastore Creation for DAE models}
For DAE models $f_{\text{DAE}}$, e.g. BART \citep{lewis-etal-2020-bart}, MBART \citep{mbart}, we generate the (key,value) pair in a way similar to vanilla \knnmtend, except that the source and target side are both target language sentences:
\begin{equation}
    (k,v) = (f_{\text{DAE}}(Y, Y_{<t}), y_t).
\end{equation}

% \paragraph{Datastore Creation for CLM Models}
For CLM models $f_{\text{CLM}}$, e.g. GPT \citep{gpt}, GPT-2 \citep{gpt2}, we generate the (key,value) pair directly using teacher forcing decoding:
\begin{equation}
    (k,v) = ( f_{\text{CLM}}(Y_{<t}), y_t).
\end{equation}

\section{Experiments}

% Please add the following required packages to your document preamble:

\subsection{Datasets}
We conduct our experiments on bilingual and multilingual machine translation benchmarks. More details of datasets and evaluation can be found in Appendix A.

\paragraph{Bilingual Machine Translation}
For bilingual machine translation, we consider low, medium, and high resource settings. For the low resource setting, we use IWSLT14 German-English dataset. For the medium resource setting, we use WMT14 English-German dataset. For the high resource setting, we use WMT17 English-Chinese dataset. 

\paragraph{Multilingual Machine Translation}
For multilingual machine translation, we use IWSLT14 multilingual dataset, which consists of parallel sentences between English and other eight languages.

\subsection{Implementation Details}

\paragraph{Tokenization}
For vanilla Transformer baselines, we learn a joint-bpe vocab using \textit{sentencepiece} \footnote{https://github.com/google/sentencepiece} toolkit. For models that utilize monolingual PLMs, we adopt the subword tokenizer on the language side that is the same as PLMs, and learn an independent subword tokenizer on the other language side. For models that utilize multilingual PLMs or MT models, we directly inherit the subword tokenizer of pre-trained models.

\paragraph{Model}
We take Transformer as the backbone of our models. For WMT datasets, we use the \textit{base} architecture following \citet{transformer}. For IWSLT datasets, we use a smaller version of Transformer. All models are implemented using \textit{fairseq} toolkit \citep{ott2019fairseq}. Details of the model hyperparameters can be found in Appendix B.

\paragraph{Optimization}
All models are optimized using Adam \citep{adam}. We train our model using a batch size of 32,000 for 100,000 steps. The learning rate is set to 5e-4. The temperature $\tau$ and balancing factor $\alpha$ are tuned based on validation performance on each dataset.

\paragraph{Datastore Creation}
We build datastores and conduct approximate nearest neighbor search using \textsc{faiss} library \citep{faiss}, and the factory template is \textit{IVFx,PQ32}. To reduce computational and storage cost, we reduce the dimension of key representations to 256 using principal component analysis before building datastores. We retrieve 8 neighbors for each query. Discussion on influences of the number of neighbors is shown in Section \ref{sec:influence_of_k}.

\subsection{Results on Bilingual Translation }

We take Transformer and \knnmt as our competitors. For PLM based datastore creation, we use XLM-R (\textit{base}), as it achieves the best performance in preliminary experiments. To make a more fair comparison, we also list the results of MT models that utilize PLMs, where the output of PLMs is fed into MT models as inputs. Previous works \citep{xu-etal-2021-bert,sun-etal-2021-multilingual-translation} show this is an effective way to incorporate PLMs. 

For MT model based datastore creation, we finetune mRASP2 \citep{pan-etal-2021-contrastive}, a recently proposed multilingual sequence-to-sequence model pre-trained for machine translation, on each dataset and treat it as the external powerful MT model. For \knnmt, we train MT models with the knowledge distillation objectives under the supervision of finetuned mRASP2 for a fair comparison.   

\begin{table}[]
\footnotesize
\centering
\begin{tabular}{l|rc|c}
 \toprule[1pt]
 &
  
  \begin{tabular}[]{@{}l@{}}Total Cost \\ (TFLOPs)\end{tabular} &
  \multicolumn{1}{c|}{\begin{tabular}[c]{@{}c@{}}Infer. Speed \\ (Tokens / s)\end{tabular}} &
  BLEU \\
  \midrule
Transformer          & 41.3  & 3411 & 34.5\\
\knnmt                & 44.0  & 2421 & 34.4\\
Transformer $^{\dagger}$  & 111.0 & 2757 & 35.3\\
\knnmt $^{\dagger}$  & 113.7 & 1905 & 35.5\\
\method $^{\dagger}$  & 45.0  & 2391 & \textbf{36.4}\\
\bottomrule[1pt]
\end{tabular}
\caption{Computational cost at the inference time, and inference speed of different models. Models with $\dagger$ have access to XLM-R as external resources.}
\label{tab:efficiency}
\end{table}

\paragraph{\method is effective.} In Table \ref{tab:main_results}, we show the experimental results on three benchmarks. Translation performances are evaluated by BLEU \citep{papineni-etal-2002-bleu} and COMET \citep{rei-etal-2020-comet}. It can be seen \method performs better than baselines in most settings. For example, when using XLM-R as the external pre-trained model, \method outperforms \knnmt by +0.8 BLEU on WMT14 En-De dataset and +0.9 BLEU on IWSLT14 De-En dataset. On the high resource setting, \method also achieves slightly better BLEU and COMET scores compared to baselines. This indicates \method is a more effective way to leverage PLMs. 

\paragraph{\method is also efficient.} We also list the computational cost (measured by FLOPs) and inference speed in Table \ref{tab:efficiency}. The details can be found in Appendix C. It can be seen that \method significantly reduces the computational cost, and slightly accelerates the inference speed compared to \knnmt with XLM-R baseline. This is because \knnmt needs to forward pass XLM-R model to get the encoding, while \method only needs to compute through the original transformer model.

\subsection{Results on Multilingual Translation}
Table \ref{tab:multilingual} lists experimental results on IWSLT14 multilingual datasets. We can see \method also achieves improvements over \knnmt on multilingual settings, when using XLM-R as external resources. It is worth mentioning that for many-to-one translation, \knnmt needs to keep multiple bilingual datastores. In contrast, \method only needs to keep one monolingual datastore of the target language, which can substantially save the storage cost.

\subsection{Comparison with Stronger Baselines}
We also compare \method to other works that aim to improve \knnmt by reducing  impact of retrieval noise \cite{zheng-etal-2021-adaptive,jiang-etal-2021-learning} and refining the representation used to build datastores \cite{wang-etal-2022-efficient,CLKNN}. Results on IWSLT14 De $\to$ En and WMT14 En $\to$ De datasets are reported in Table \ref{fig:strong_baseline}, where all models have access to XLM-R as external resources. It can be seen that \method still outperforms these baselines by a large margin, validating the effectiveness of \methodend.

\begin{table}[t]
\footnotesize
\centering
\begin{tabular}{l|c|c}
\toprule[1pt]
                                             & En $\to$ X & X $\to$ En \\
                                             \midrule
Transformer                                  & 25.5    & 28.9                   \\
\knnmt                                       & 25.7    & 29.0\\
Transformer $^{\dagger}$             & 25.7    & 29.2                   \\
\knnmt$^{\dagger}$                  & 25.9       & 29.6                   \\
\method$^{\dagger}$               & \textbf{26.4}       & \textbf{29.9}                  \\
\bottomrule[1pt]
\end{tabular}
\caption{Multilingual translation results on IWSLT14 datasets. Average BLEU scores on seven languages are reported. Models with $\dagger$ have access to XLM-R as external resources.}
\label{tab:multilingual}
\end{table}

\begin{table}[t]
\footnotesize
\centering
\begin{tabular}{l|cc}
\toprule
               & IWSLT14 De En & WMT14 En De \\
               \midrule
Transformer    & 35.4          & 27.8        \\
\knnmt         & 35.7          & 27.9        \\
Adaptive \knnmt & 35.8          & 28.0        \\
KSTER          & 35.8          & 28.2        \\
PCKMT          & 35.9          & 28.2        \\
CLKNN         & 35.7          & 28.3        \\
\method           & \textbf{36.4}          & \textbf{28.7}       \\
\bottomrule
\end{tabular}
\caption{Comparison of \method and other methods that aims to improve \knnmtend, including Adaptive \knnmt \citep{zheng-etal-2021-adaptive}, KSTER \citep{jiang-etal-2021-learning}, PCKMT \citep{wang-etal-2022-efficient} and CLKNN \citep{CLKNN}. All models have access to XLM-R as external resources.}
\label{fig:strong_baseline}
\end{table}

\begin{figure}[t]
    \centering
    \includegraphics[width=0.8\linewidth]{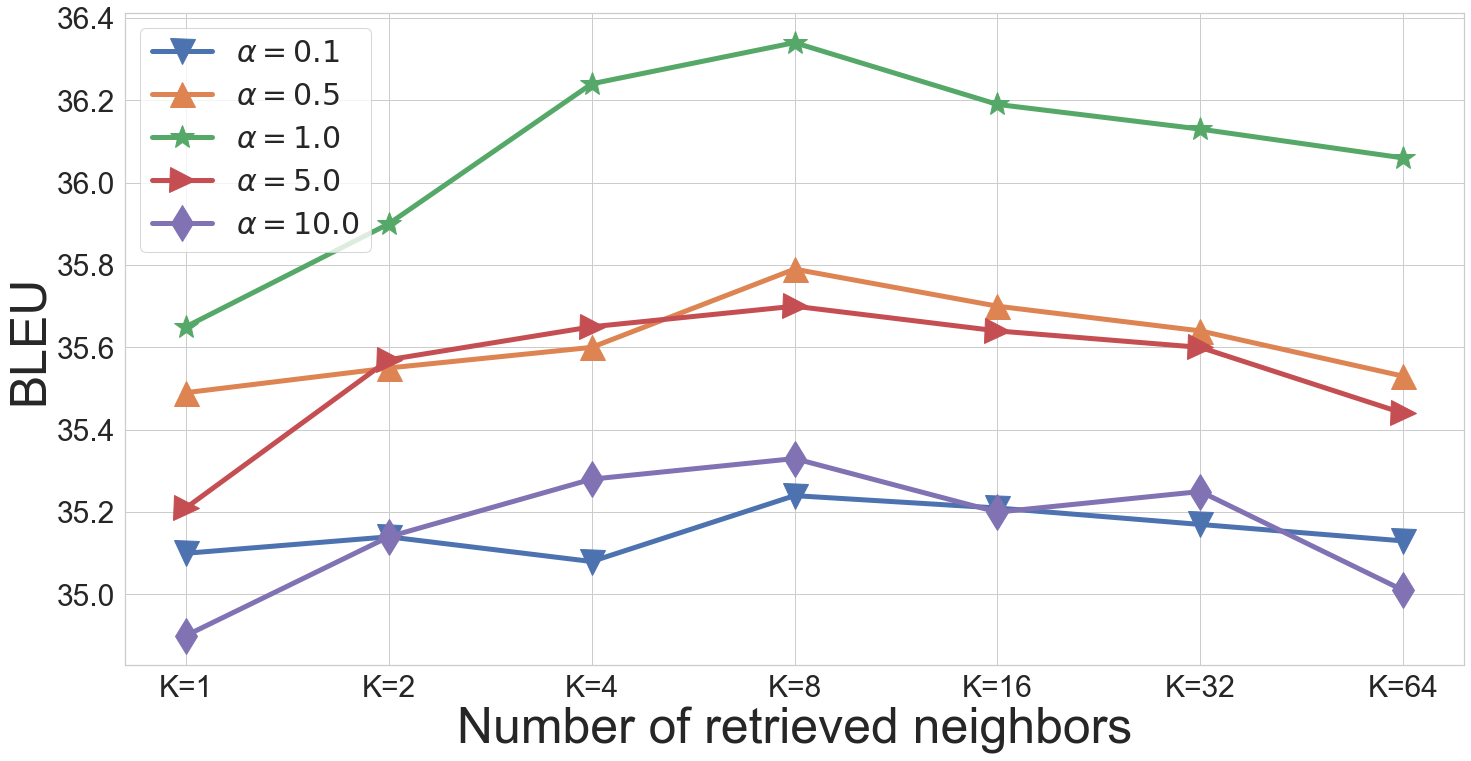}
    \caption{Influence of $k$ and $\alpha$ on IWSLT14 De $\to$ En dataset.}
    \label{fig:influence_of_k}
\end{figure}

\section{Discussion}

\begin{figure*}[htbp]
    \centering
    \begin{subfigure}[b]{0.48\textwidth}
        \includegraphics[width=\textwidth]{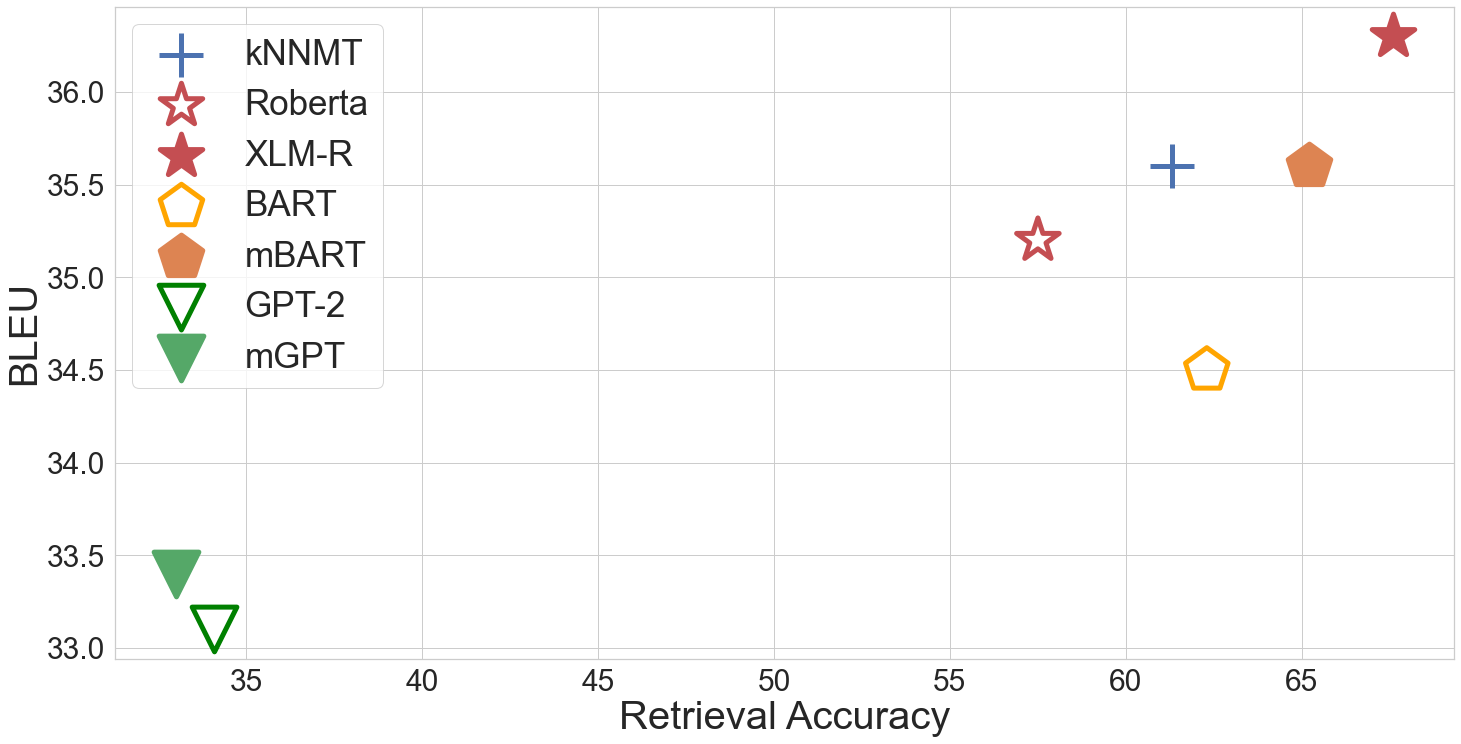}
        \caption{}
        \label{fig:plm_bleu}
    \end{subfigure}
    % \hspace{1em}
    \begin{subfigure}[b]{0.48\textwidth}
        \includegraphics[width=\textwidth]{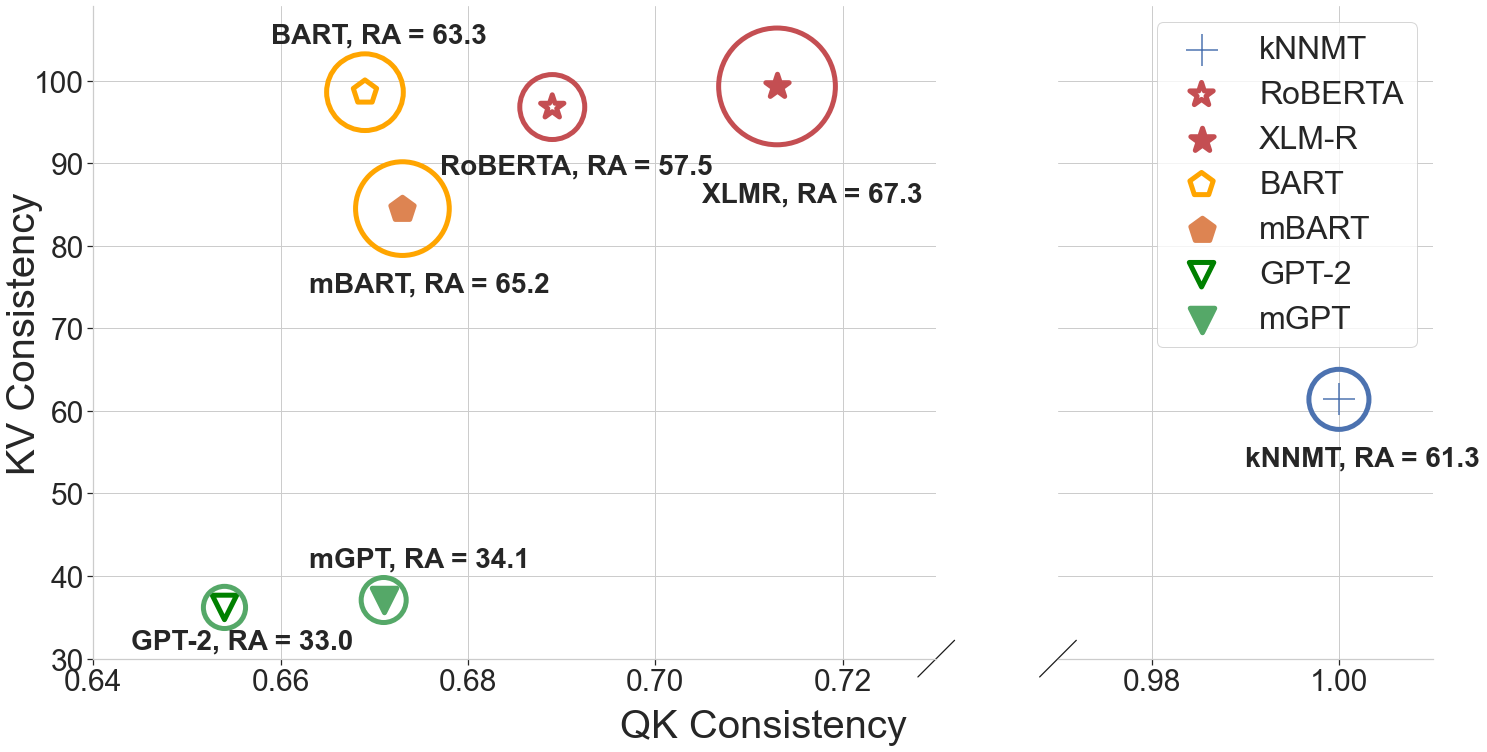}
        \caption{}
        \label{fig:qk_kv}
    \end{subfigure}
    \caption{(a) Retrieval accuracy and BLEU score  (b) Q-K consistency, K-V consistency, and retrieval accuracy of \knnmt augmented by XLM-R and \method augmented by various PLMs. In (b), larger circle around the data point means higher retrieval accuracy (RA).} %\shanbo{这个图和最开始的图一样，线条和字都太小太浅了，如果打印出来的话基本上很难看得清楚。可以线条颜色深一点，字大一点 or 粗一点}}
    \label{fig:plm}
\end{figure*}

\subsection{Influence of $k$ and $\alpha$}
\label{sec:influence_of_k}
We vary the  number of retrieval neighbors  $k$ and balancing factor $\alpha$, and evaluate the corresponding translation performance of \method on IWSLT14 De $\to$ En dataset. Results are shown in Figure \ref{fig:influence_of_k}. We notice that retrieving too many neighbors does not lead to better translation performance. We hypothesize larger $k$ would incur more noise in the retrieval results. Empirically, we find setting $k = 8$ and $\alpha = 1.0$ is a good choice.

\subsection{Effectiveness of Contrastive Training }
\label{sec:mse_vs_contrast}
For the alignment objective, it is also possible to directly minimize the mean squared error (MSE) distance between queries and keys as in knowledge distillation \citep{kd} and previous work \citep{zheng-etal-2021-non-parametric}. We compare MSE with the proposed NCA objective in Table \ref{tab:mse_vs_nca}. As can be seen, the NCA objective surpasses the MSE objective by a large margin across four translation directions. 

To better understand this phenomenon, we also provide visualization of the representations of queries and keys using bivariate kernel density estimation in Figure \ref{fig:kde}. Experimental details can be found in Appendix D. It can be seen that training with the NCA objective could lead to more aligned representations between queries and keys.

\begin{table}[tbp]
\footnotesize
\centering
\begin{tabular}{l|c|c|c|c}
\toprule[1pt]
    & \multicolumn{2}{c|}{IWSLT14} & \multicolumn{2}{c}{WMT14} \\ \midrule
    & De-En        & En-De        & De-En       & En-De       \\ \midrule
MSE & 34.9        & 28.8       &   32.5           &   28.1           \\
%CA  & 35.6        & 29.4       &   32.7           &  28.3           \\ 
NCA & \textbf{36.4}       & \textbf{30.1}     &      \textbf{33.0}     & \textbf{28.7}     \\
\bottomrule[1pt]
\end{tabular}
\caption{Comparison on BLEU of different aligning methods.}
\label{tab:mse_vs_nca}
\end{table}

\begin{figure}[t]
    \centering
    \begin{subfigure}[b]{0.22\textwidth}
        \includegraphics[width=\textwidth]{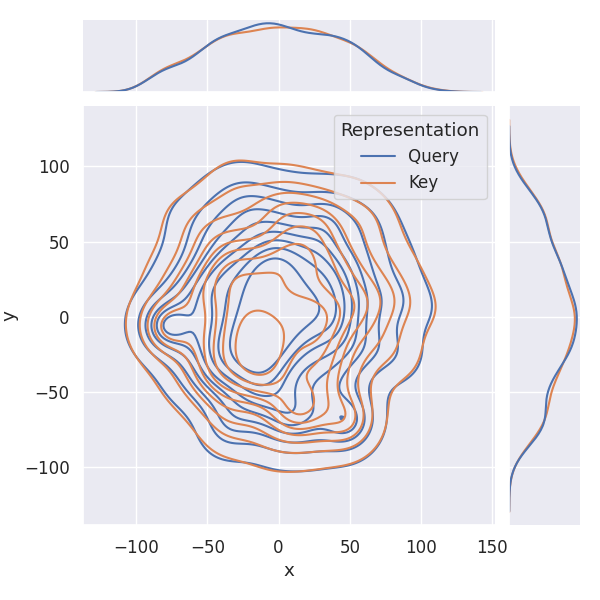}
        \caption{MSE}
        \label{fig:mse}
    \end{subfigure}
    \begin{subfigure}[b]{0.22\textwidth}
        \includegraphics[width=\textwidth]{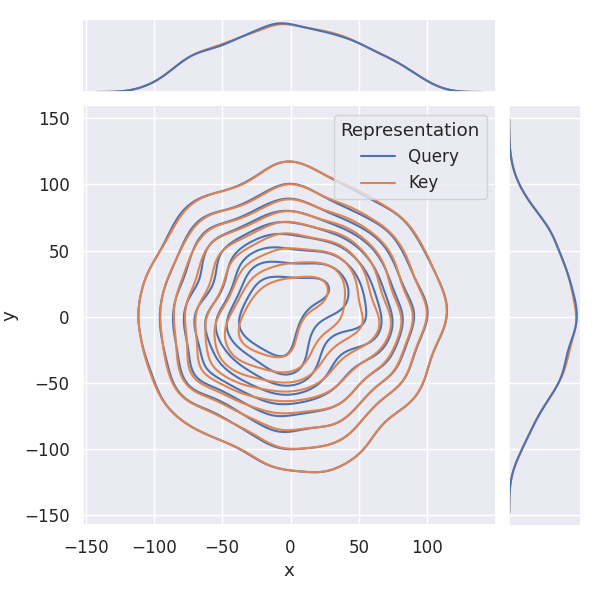}
        \caption{NCA}
        \label{fig:nca}
    \end{subfigure}
    \caption{Bivariate KDE of query and key representation from models trained with MSE  and NCA  objectives.}
    \label{fig:kde}

\end{figure}

\subsection{Impact of Choices of PLMs }

We have shown \method brings substantial improvements over baselines with the help of XLM-R, a multilingual PLM pre-trained with MLM objective. In this section, we present a further study on the relationship between choices of PLMs and \methodend's translation performance.

Concretely, we investigate two factors that may affect PLMs' performance when used to build datastores: training corpus and training objective. We consider monolingual/multilingual settings for the former, and MLM/DAE/CLM for the latter. For each combination of factors, we select a representative pre-trained model, summing up to 6 models: RoBERTA (monolingual MLM), XLM-R (multilingual MLM), BART (monolingual DAE), mBART (multilingual DAE), GPT-2 (monolingual CLM), and mGPT (multilingual CLM). We build a datastore from each model on IWSLT14 De $\to$ En dataset, and train \method to retrieve from these datastores. We plot the translation performance, retrieval accuracy \footnote{The computation of retrieval accuracy is similar to Equation \ref{eqn:kv_cons}, except that queries come from the NMT model instead of the datastore.}, Q-K consistency and K-V consistency in Figure \ref{fig:plm}.

\paragraph{(Appropriate) PLMs provide better datastores.} From Figure \ref{fig:qk_kv}, we can see that datastores from PLMs (except GPT-2 and mGPT, which we will discuss later) have higher K-V consistency compared to \knnmtend. This validates our motivation of leveraging PLMs to build better datastores.

\paragraph{Multilingual pretraining is more beneficial than monolingual pretraining.}  In Figure \ref{fig:plm_bleu}, multilingual pre-trained models show consistent advantages over their monolingual counterparts. We hypothesize the reason is that machine translation is a cross-lingual task, so aligning the MT model's representations to multilingual PLMs' takes less effort than to monolingual PLMs'. This is  confirmed in Figure \ref{fig:qk_kv}: \method with multilingual PLMs enjoy higher Q-K consistency than with those with monolingual PLMs.

\paragraph{MLM objective is better than DAE and CLM objective.}
Another interesting trend is that models trained with MLM datastores perform the best among the three training objectives, while models trained with CLM datastores perform poorly.  From Figure \ref{fig:qk_kv}, we can see that K-V consistency of CLM models is low compared to MLM and DAE models, which leads to low retrieval accuracy of the final model, hindering the translation performance. 
This is because the information about target sentences that CLM models possess is incomplete due to their unconditional auto-regressive nature, making them unable to align their representations $f_{CLM}(Y_{<t})$ with the corresponding $y_t$ very well.
In contrast,  the whole target sentence is explicitly fed to encoders in MLM and DAE models, thus substantially reducing the uncertainty. 

Comparing MLM and DAE objectives, we can see although datastores built from MLM and DAE models achieve similar K-V consistency, Q-K consistency of models trained with DAE datastores is relatively lower than models with MLM datastores, leading to lower retrieval accuracy. This indicates it is more difficult to align MT representations to DAE models' representations than to MLM models' representations. We leave the investigation of reasons to future works.

\subsection{Case Study}

We present translation examples of baseline model and \method in Table \ref {tab:case_study}. We also list the retrieval neighbors at the timestep where models generate incorrect words. 
In the first example, we can see  although the original Transformer model generates the correct word ``\textit{friend}", the word \knnmt generates (``\textit{liberated}") is incorrect due to its low-quality retrieval results. In contrast,  \method mitigates this negative effect by retrieving more accurate results.

We can also observe that \method is better at distinguishing similar but not exchangeable words, e.g., personal pronouns in the second example. We take the idea of contrastive translation evaluation \citep{rios-gonzales-etal-2017-improving} to quantify this phenomenon better. More details can be seen in Appendix E. Results are shown in Figure \ref{fig:contrastive_evaluation}.  It can be seen that \method consistently outperforms \knnmt on all word categories, demonstrating the effectiveness of the contrastive training objective.

\begin{table}[t]
\footnotesize
\centering
\begin{tabular}{c|l}
\toprule
SRC             & eine befreundete journalistin hatte ... .                                    \\
REF       & a journalist \textit{friend} has been ....                                            \\\midrule
MT              & a \textit{friend} journalist has been ... .                                            \\
\knnmt           & a \underline{liberated} journalist had been ...   \\
\method             & a \textit{friend} journalist had been ... \\ \midrule
\knnmt Re. & \underline{liberated liberated} \textit{friend} \underline{liberated} \\
\method Re.  & \textit{friend friend friend friend}  \\ \bottomrule  
% \toprule
% Src             & wenn dan eine marke wäre ...                         \\
% Reference       & if dan \green{were} a brand, ...                                            \\\midrule
% MT              & if dan \red{was} a brand, ...                                            \\
% \knnmt          & if dan \red{was} a brand, ...   \\
% \method            & if dan \green{were} a brand, ...\\ \midrule
% \knnmt Retr. & \red{was was found was}      \\
% \method Retr.  & \green{were were were were}   \\ \bottomrule 
\toprule
SRC            & sie haben ihre bilder mitgebracht, ja?                                   \\
REF       & \textit{you} brought your pictures, right?                                      \\\midrule
MT              & \underline{they} brought their pictures with me, right?                                            \\
\knnmt           & \underline{they} brought their pictures with me, right?   \\
\method             & \textit{you} brought your pictures with you, right? \\ \midrule
\knnmt Re. & \underline{they they} \textit{you} \underline{they} \\
\method Re. & \textit{you you you you}   \\ \bottomrule 
\end{tabular}
\caption{Translation and retrieval examples of \knnmt and \methodend. Re. means the retrieved top 4 examples. Ground truth words and correct retrieved words are in italic, while incorrect translated and retrieved words are underlined.}
\label{tab:case_study}
\end{table}

\section{Related Work}
\subsection{Retrieval-Enhanced Machine Translation}
RE-NMT, which enhances neural machine translation systems with retrieval mechanism, has been shown effective in many previous works. Most RE-NMT methods are based on sentence or $n$-gram level retrieval \citep{zhang-etal-2018-guiding,gu2018search,xia2019graph,he-etal-2021-fast,cai-etal-2021-neural}. Our work differs from theirs in that we retrieve at the token level, which significantly eases the data sparsity problem and enables our method to retrieve more relevant examples. 

\begin{figure}[t]
    \centering
    \includegraphics[width=0.7\linewidth]{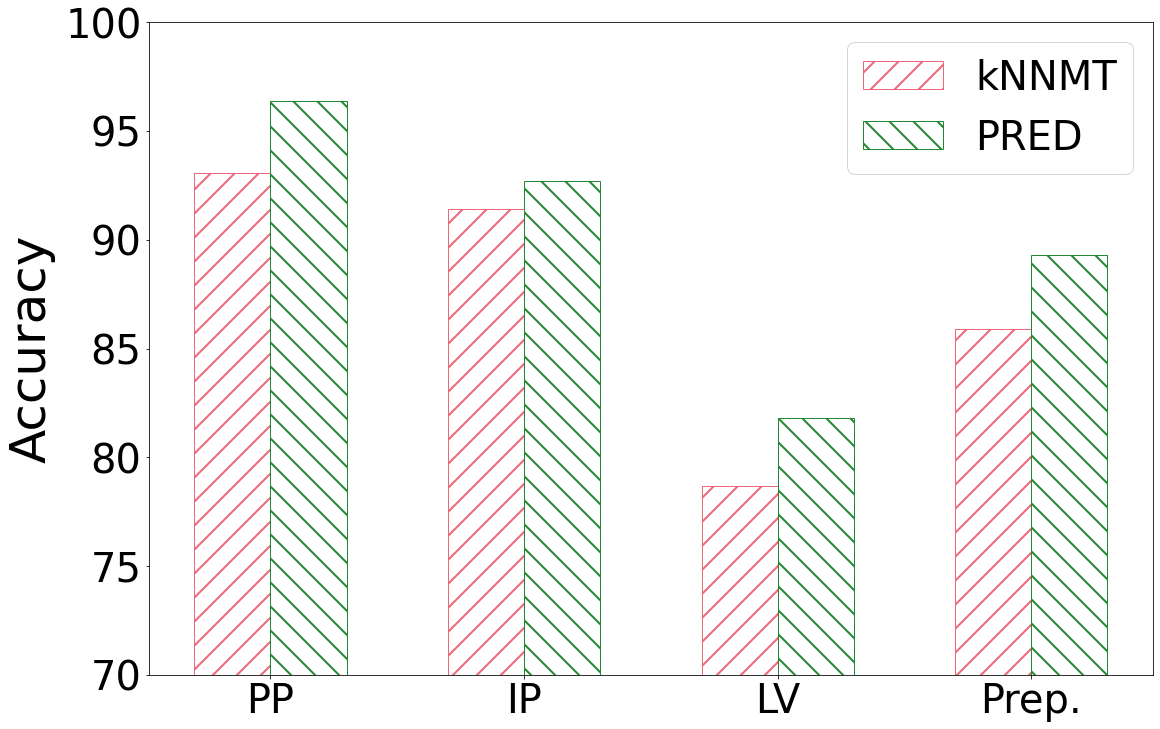}
    \caption{Accuracy of contrastive evaluation on different kinds of words. PP: personal pronouns. IP: indicative pronouns. LV: linking verbs. Prep.: prepositions.}
    \label{fig:contrastive_evaluation}
\end{figure}

There is also a line of research that retrieves at the token level.
\citet{knnmt} firstly propose to retrieve similar tokens based on the similarity of decoding representations. \citet{zheng-etal-2021-adaptive} learn a Meta-$k$ network to select the optimal $k$ at different timestep. \citet{jiang-etal-2021-learning} propose to smooth the distribution of retrieved results and learn the interpolation weight automatically. \citet{fast_knnmt} propose to prune the datastore firstly to accelerate the retrieval process. All these works build their datastores using the original MT representation, while we explore ways to leverage powerful pre-trained models for datastore creation.

The most relevant work to ours is \citet{zheng-etal-2021-non-parametric}, which adapts the original MT model to generate monolingual datastores. However, the datastore generator in their work is still learned from the limited size of MT corpus, thus facing the same representation quality problem as \knnmtend.

\subsection{Pre-trained Language Models for Neural Machine Translation}
Many works attempt to make use of large-scale pre-trained language models for NMT. \citet{xlm,conneau-etal-2020-unsupervised,mbart,xu-etal-2021-bert}  initialize NMT models with parameters of multilingual PLMs, and achieve substantial improvement in both supervised and unsupervised machine translation. \citet{zhu2019incorporating,yang2020towards} fuse BERT's representation to NMT models in a dynamic way. \citet{chen-etal-2020-distilling} first finetune BERT to be a conditional MLM model, and distill the knowledge of it to NMT models. Unlike previous works, PLMs in our work are used to create better datastores.
\section{Conclusion}
We introduce a framework that generates better datastores in \knnmtend. Pre-trained models can be efficiently leveraged to create datastores, and a unified model that conducts retrieval and translation is trained via a novel contrastive objective. Experiments demonstrate the effectiveness of our method. Our work opens the gate to making use of pre-trained models for RE-NMT. We believe investigating how to better create datastores from PLMs, and how to leverage more monolingual corpus to benefit \method are promising future directions.
\section*{Acknowledgement}
Shujian Huang is the corresponding author. This work is supported by National Science Foundation of China (No. 6217020152), the Liaoning Provincial Research Foundation for Basic Research.
% Entries for the entire Anthology, followed by custom entries
\bibliography{anthology,custom}

\appendix
\section{Details of Datasets and Evaluation}
\paragraph{Bilingual translation datasets} We use WMT17 English-Chinese, WMT14 English-German, IWSLT14 German-English datasets for bilingual machine translation. The statistics of each dataset are listed in Table \ref{tab:bilingual_dataset_statistics}.

\begin{table}[t]
\centering
\begin{tabular}{c|ccc}
\toprule
              & Train   & Valid & Test \\
              \midrule
WMT17 En-Zh   & 19481469 & 2002  & 2001 \\
WMT14 En-De   & 4590101  & 3000  & 3000 \\
IWSLT14 En-De & 160239  & 7283  & 6750 \\
\bottomrule
\end{tabular}
\caption{ Statistics of bilingual translation datasets used in this paper.}
\label{tab:bilingual_dataset_statistics}
\end{table}

\paragraph{Multilingual translation dataset} We use IWSLT14 multilingual dataset for multilingual translation experiments. The dataset includes parallel sentences of English to Arabic (ar), German (de), Spanish (es), Persian (fa), Hebrew (he), Italian (it), Dutch (nl) and Polish (pl). The number of parallel sentences for each language pair is shown in Table \ref{tab:multilingual_dataset_stats}.

\begin{table}[t]
\centering
\begin{tabular}{c|ccc}
\toprule
   & Train  & Valid & Test \\
   \midrule
ar & 139748 & 6352  & 5357 \\
de & 160239 & 7283  & 5585 \\
es & 169028 & 7683  & 5593 \\
fa & 89230  & 4055  & 4244 \\
he & 144345 & 6561  & 5594 \\
it & 167195 & 7599  & 5553 \\
nl & 153590 & 6981  & 5389 \\
pl & 128410 & 5836  & 5462 \\
\bottomrule
\end{tabular}
\caption{ Statistics of multilingual translation datasets used in this paper.}
\label{tab:multilingual_dataset_stats}
\end{table}

\paragraph{Automatic Evaluation Metrics} We use BLEU \citep{papineni-etal-2002-bleu} and COMET \citep{rei-etal-2020-comet} to evaluation the translation performance in this paper. For IWSLT14 datasets, we use a beam of 5, while for other datasets we use a beam of 4. We compute the COMET score using \textit{wmt20-comet-da} model provided by the official COMET repository\footnote{https://github.com/Unbabel/COMET}.

\section{Model Hyperparameters}
We use the \textit{base} and \textit{small} configurations for Transformer in this paper. The hyperparameters of each configuration are shown in Table \ref{tab:hyperparameters}.

\begin{table}[]
\centering
\begin{tabular}{c|cc}
\toprule
                & \textit{base} & \textit{small} \\
                \midrule
Encoder Layer   & 6    & 6     \\
Decoder Layer   & 6    & 6     \\
embedding dim   & 512  & 512   \\
hidden dim      & 2048 & 1024  \\
attention heads & 8    & 4     \\
dropout         & 0.1  & 0.3  \\
\bottomrule
\end{tabular}
\caption{Hyperparameters of the model configuration used in this paper.}
\label{tab:hyperparameters}
\end{table}

\section{Details of The Computation of FLOPs and Inference Speed}
% \paragraph{Number of parameters} We use the number reported by \textit{fairseq} toolkit, and the snippet used to count the parameters can be found in the repository\footnote{\url{https://github.com/facebookresearch/fairseq/blob/acd9a53607d1e5c64604e88fc9601d0ee56fd6f1/fairseq_cli/train.py\#L115}}.

\paragraph{Model FLOPs} The FLOPs of model forward is computed using scripts provided by \citet{clark2020electra}\footnote{\url{https://github.com/google-research/electra/blob/master/flops_computation.py}}, and the FLOPs of retrieval process is manually computed following \citet{jegou2010product}.  FLOPS reported in the paper refers to the total FLOPs needed to inference over the test set of IWSLT14 De $\to$ En dataset.

\paragraph{Inference Speed} The hardware environment used to test the inference speed is Intel(R) Xeon(R) Platinum 8260 CPU @ 2.40GHz + 1 NVIDIA V100 GPU. Average inference speed is reported using a batch size of 128 sentences.

\section{Details of Visualization of Query and Key Representation}
To better understand the superiority of NCA object, we visualize the degree of alignment between the representation space of queries and keys in models trained with MSE and NCA objectives, respectively. Specifically, we collect 10,000 queries and key representations from each model on IWSLT 14 De $\to$ En dataset and reduce their dimension to 2 using t-SNE. The degree of alignment is depicted by bivariate kernel density estimation.

\section{Details of Contrastive Evaluation}
A contrastive translation pair \citep{rios-gonzales-etal-2017-improving} contains a source, a reference, and one or more contrastive translations. Contrastive translations are constructed by substituting words in the reference according to specific rules. NMT systems are used to score reference and contrastive translations. If the reference is scored higher than all contrastive translations, then the NMT system passes the contrastive translation test.

Specifically, we consider four kinds of words for substitution: person pronouns (\textit{I/you/he/she}), indicative pronouns (\textit{this/that/these/those}), linking verbs (\textit{is/are/was/were}) and prepositions (\textit{in/on/at}). The accuracy of the contrastive translation test of \knnmt and \method are reported.

\end{document}